\documentclass{article} 
\usepackage{iclr2016_conference,times}
\usepackage{hyperref}
\usepackage{url}

\usepackage{siunitx}
\usepackage{floatrow}
\newcommand{\Isz}[2]{I_{\scriptscriptstyle {#1}\times{#2}}}

\usepackage[]{algorithm2e}

\title{Black box variational inference for state space models}

\author{%
Evan Archer \\
Department of Statistics and Grossman Center\\
Columbia University \\
New York City, NY, United States \\
\texttt{evan@stat.columbia.edu}
\And
Il Memming Park \\
Department of Neurobiology and Behavior\\
Stony Brook University \\
Stony Brook, NY, United States \\
\texttt{memming.park@stonybrook.edu}
\And
Lars Buesing\thanks{Current affiliation: Google DeepMind, London, UK.} \\
Department of Statistics and Grossman Center \\
Columbia University\\
New York City, NY, United States \\
\texttt{lbuesing@google.com}
\AND
John Cunningham \\
Department of Statistics and Grossman Center\\
Columbia University \\
New York City, NY, United States \\
\texttt{jpc2181@columbia.edu}
\And
Liam Paninski \\
Department of Statistics and Grossman Center\\
Columbia University \\
New York City, NY, United States \\
\texttt{liam@stat.columbia.edu}
}

\usepackage{bm}  
\usepackage{mathrsfs}
\usepackage[pdftex]{graphicx}
\usepackage{amsmath,amssymb,amsfonts,amsthm} 
\usepackage{sidecap}  
\sidecaptionvpos{figure}{t}

\newcommand{\mybibfile}{bibliography}

\usepackage{mathtools} 

\newcommand{\optionrule}{\noindent\rule{1.0\textwidth}{0.75pt}}

\renewcommand{\eqref}[1]{eq.~\ref{eq:#1}}
\newcommand{\figref}[1]{Fig.~\ref{fig:#1}}

\newcommand{\inv}[1]{\ensuremath{{#1}^{-1}}} 

\newcommand{\R}{\mathbb R}

\newcommand{\trp}{^{\mathrm{T}}} 
\newcommand{\itrp}{^{-\mathrm{T}}} 

\newcommand{\E}{\mathbb{E}}
\newcommand{\V}{\mathbb{V}}

\usepackage{forloop}
\newcommand{\defvec}[1]{\expandafter\newcommand\csname v#1\endcsname{{\mathbf{#1}}}}
\newcounter{ct}
\forLoop{1}{26}{ct}{
    \edef\letter{\alph{ct}}
    \expandafter\defvec\letter
}
\newcommand{\defmat}[1]{\expandafter\newcommand\csname m#1\endcsname{{\mathbf{#1}}}}
\forLoop{1}{26}{ct}{
    \edef\letter{\Alph{ct}}
    \expandafter\defmat\letter
}

\newcommand{\vepsilon}{\bm{\epsilon}}

\newcommand{\grad}{\nabla}

  \newtheoremstyle{evandefinition}{\topsep}{\topsep}%
     {}
     {}
     {\bfseries}
     {}
     {\newline}
     {\thmname{#1}\thmnumber{ #2}. \textit{\thmnote{ #3} }}

     \newtheoremstyle{indenteddefinition}{\topsep}{\topsep}
     {\addtolength{\leftskip}{2em}} 
     {-1.75em}
     {\bfseries}
     {}
     { }
     {\thmname{#1} \thmnumber{#2}. \textbf{(\thmnote{#3})}}

\theoremstyle{evandefinition}

\theoremstyle{indenteddefinition}

\theoremstyle{remark}

\makeatletter
\renewcommand*\env@matrix[1][c]{\hskip -\arraycolsep
  \let\@ifnextchar\new@ifnextchar
  \array{*\c@MaxMatrixCols #1}}
\makeatother

\begin{document}

\maketitle

\begin{abstract}
Latent variable time-series models are among the most heavily used tools from machine learning and applied statistics. These models have the advantage of learning latent structure both from noisy observations and from the temporal ordering in the data, where it is assumed that meaningful correlation structure exists across time.  A few highly-structured models, such as the linear dynamical system with linear-Gaussian observations, have closed-form inference procedures (e.g. the Kalman Filter), but this case is an exception to the general rule that exact posterior inference in more complex generative models is intractable.  Consequently, much work in time-series modeling focuses on approximate inference procedures for one particular class of models. Here, we extend recent developments in stochastic variational inference to develop a `black-box' approximate inference technique for latent variable models with latent dynamical structure.  We propose a structured Gaussian variational approximate posterior that carries the same intuition as the standard Kalman filter-smoother but, importantly, permits us to use the same inference approach to approximate the posterior of much more general, nonlinear latent variable generative models. We show that our approach recovers accurate estimates in the case of basic models with closed-form posteriors, and more interestingly performs well in comparison to variational approaches that were designed in a bespoke fashion for specific non-conjugate models.
\end{abstract}

\section{Introduction}

Latent variable models are commonplace in time-series analysis, with applications across statistics, engineering, the sciences, finance and economics. The core approach is to assume that latent variables $\vz_1, \dots, \vz_T\in \R^n$, which are correlated across $t$, underlie correlated observations $\vx_1, \dots, \vx_T\in\R^m$. Standard models for latent dynamics include the linear dynamical system (LDS) and hidden Markov models. While each approach comes with a distinct model and set of computational tools, often the basic goal of inference is the same: to discern the filtering distribution $p(\vz_t | \vx_{1:t})$ and the smoothing distribution $p(\vz_t | \vx_{1:T})$ of the latent variables. Closed-form expressions for these distributions are available when the overall probabilistic model has tree or chain structure that admits closed-form message passing.  In general, inference in non-Gaussian or nonlinear models requires numerical approximation or sampling.

Markov chain Monte Carlo sampling and particle filtering for general time-series models are well-developed but typically do not scale well to large-scale problems. Even when a model $p_\theta(\vx,\vz)$, with parameters $\theta$, has been trained, we can only access an analytically-intractable posterior through further sampling. Here, we take a variational approach to time-series modeling: rather than attempting to compute the posterior $p_\theta (\vz|\vx)$ of our generative model, we approximate it with a distribution $q_\phi(\vz|\vx)$ with variational parameters $\phi$. Inference proceeds by simultaneously optimizing $p_\theta$ (through its model parameters $\theta$)  and $q_\phi$ (through its variational parameters $\phi$) such that $q_\phi$ approximates the true posterior.

Our main contributions are (1) a structured approximate posterior that can express temporal dependencies, and (2) a fast and scalable inference algorithm. We propose a multivariate Gaussian approximate posterior with block tri-diagonal inverse covariance, and formulate an algorithm that scales (in both time and space complexity) only linearly in the length of the time-series. For inference, we make use of recent advances in stochastic gradient variational Bayes (SGVB) \citep{Rezende2014, Kingma2013, Kingma2014} to learn an approximate posterior with a complex functional dependence upon the observations $\vx$. Using this approach we are able to learn a neural network (NN) that maps $\vx$ into the smoothed posterior $q(\vz|\vx)$ (sometimes called the recognition model). This approach is also `black-box' in the sense that the inference algorithm does not depend explicitly upon the functional form of the generative model $p_\theta$.

Our motivations lie in the study of high-dimensional time-series, such as neural spike-train recordings~\citep{Kao2015b}. We seek to infer trajectories $\vz$ that provide insight into the latent, low-dimensional structure in the dynamics of such data. Recent, related approaches to variational inference in time-series models focus upon the design and learning of rich generative models capable of capturing the statistical structure of large, complex datasets \citep{Gan2015, Chung2015}. In contrast, our focus is upon computationally efficient inference in structured, interpretable parameterizations that build upon methods fundamental in scientific applications.

We apply our smoothing approach for approximate posterior inference of a well-studied generative model: the Poisson linear dynamical system (PLDS) model. We find that our general, black-box approach outperforms a specialized variational Bayes expectation maximization (VBEM) approach \citep{Kahn2013} to inference in PLDS, reaching comparable solutions to VBEM before it can complete a single EM iteration. Additionally, we apply our method to inference in a one-dimensional, nonlinear dynamical system, showing that we are able to accurately recover nonlinear relationships in the posterior mean. 

\section{Stochastic gradient variational Bayes}\label{sec:variational_background}

In variational inference we approximate an intractable posterior distribution $p_\theta(\vz|\vx) = p_\theta(\vx,\vz)/p_\theta(\vx)$ with $q_\phi(\vz|\vx)$\footnote{In many approaches to variational inference the dependence of upon $\vx$ is dropped; in our case, the parameterization of $q_\phi(\vz|\vx)$ may depend explicitly upon the observations $\vx$.} that comes from a tractable class (e.g., the Gaussian family) and is parameterized by variational parameters $\phi$. We learn $\phi$ and $\theta$ together by optimizing the \textit{evidence lower bound} (ELBO) of the marginal likelihood \citep{Jordan1999}, given by,
\begin{align}
  \log p_\theta(\vx) \geq \mathcal{L}(\theta, \phi ; \vx) &= \E_{q_\phi(\vz|\vx)}\left[-\log q_\phi(\vz|\vx) + \log p_\theta(\vx,\vz) \right]
\\
                                                          &= H(q_\phi(\vz|\vx)) +  \E_{q_\phi(\vz|\vx)}\left[\log p_\theta(\vx,\vz) \right]. \label{eq:lower_bound}
\end{align}

The quantity $\mathcal{L}(\theta,\phi;\vx)$ is the ELBO, and $H(q_\phi(\vz|\vx))$ is the entropy of the approximating posterior.  Our goal is to differentiate $\mathcal{L}(\theta, \phi)$ with respect to $\phi$ and $\theta$ so as to maximize $\mathcal{L}$,
\begin{align}\label{eq:grad_lower_bound}
  \grad \mathcal{L}(\theta, \phi;\vx) := \grad \E_{q_\phi(\vz|\vx)}[\underbrace{-\log q_\phi(\vz|\vx) + \log p_\theta(\vx,\vz)}_{:=f_{\left\{\theta, \phi\right\}}(\vz)}].
\end{align}

For the remainder of this section, we use the notation $f_{\left\{\theta, \phi\right\}}(\vz) = -\log q_\phi(\vz|\vx) + \log p_\theta(\vx,\vz)$. Typically at least some terms of \eqref{grad_lower_bound} cannot be integrated in closed form. While it is often possible to estimate the gradient by sampling directly from $q(\vz|\vx)$, in general the approximate gradient exhibits high variance \citep{Paisley2012}. One approach to addressing this difficulty, independently proposed by \cite{Kingma2013}, \cite{Rezende2014} and \cite{Titsias2014}, is to compute the integral using the ``reparameterization trick'': choose an easy-to-sample random variable $\epsilon$ with distribution $p(\epsilon)$ and parameterize $\vz$ through a function $g$ of observations $\vx$ and parameters $\phi$,
\begin{align}
  \vz = g_\phi(\vx, \epsilon).
\end{align}

The point of this notation is to make clear that $g_\phi(\vx, \cdot)$ is a \textit{deterministic} function: all randomness in $q$ comes from the random variable $\epsilon$. This allows us to approximate the gradient using the simple estimator,
\begin{align}
  \label{eq:sgvb_estimator}
  \grad\E_{q_\phi(\vz)}\left[f_{\left\{\theta, \phi\right\}}(\vz)\right] =  \E_{p(\epsilon)}\left[ \grad f_{\left\{\theta, \phi\right\}}(g_\phi(\vx,\epsilon))\right] \approx \frac{1}{L} \sum_{l=1}^L \grad f_{\left\{\theta, \phi\right\}}(g_\phi(\vx, \epsilon^l)),
\end{align}
where $\epsilon^l$ are iid samples from $p(\epsilon)$. In \cite{Kingma2013} this estimator is referred to as the Stochastic Gradient Variational Bayes (SGVB) estimator. Empirically, \eqref{sgvb_estimator} has much lower variance than previous sampling-based approaches to the estimation of \eqref{grad_lower_bound} \citep{Kingma2013, Titsias2014}.

An important property of \eqref{sgvb_estimator} is that it does not depend upon the particular form of $f_{\left\{\theta, \phi\right\}}(\cdot)$: we need only be able to evaluate it at the samples $\epsilon^i$. It is in this sense that our approach is ``black-box'': in principle, inference works the same way regardless of our choice of generative model $p_\theta$. In practice, of course, different modeling choices will affect the computation time and the convergence rate of the method.

The estimator also permits significant freedom in our parameterization of the transformation $g_\phi(\vx, \cdot)$: for inference, we just need to be able to differentiate $g$ with respect to $\phi$. While it is possible to use the SGVB approach with a separate set of parameters $\phi_t$ for each observation (as in \cite{Hoffman2013}, for instance), much recent work has used deep neural networks (DNNs) to train a function that maps directly into the posterior \citep{Rezende2014, Kingma2013, Kingma2014}. Under this approach, with a trained $g_\phi(\vx, \cdot)$, no additional gradient steps are needed to obtain $q(\vz|\vx)$ for new observations $\vx$.

\section{Variational approach to state-space modeling}

Using a black-box inference approach, learning a state-space model is in part just a matter of parameterizing a generative model $p_\theta$ with time-series structure. However, the posterior $p(\vz|\vx)$ will in general have temporal correlation structure inadequately captured by the approximate posteriors studied in most previous variational inference literature. The first challenge, then, is to formulate an approximate posterior expressive enough to capture the temporal correlations characteristic of time-series models. We take $\vx$ and $\vz$ as ``stacked'' versions of the response and latent variable, respectively, at a particular time $t$. We let $\vx = \left(\vx_1, \dots, \vx_t, \dots, \vx_T\right)$ and $\vz = \left(\vz_1, \dots, \vz_t, \dots, \vz_T\right)$, where $\vx_t\in\R^{m}$ and $\vz_t\in \R^{n}$.

\subsection{Gaussian approximate posterior}
One common, convenient choice of approximate posterior is the multivariate normal. In the notation of Section \ref{sec:variational_background}, the multivariate normal comes about if we choose $\epsilon \sim \mathcal{N}(0,I)$ and take $g_\phi(\vx, \cdot)$ to be an affine function \citep{Titsias2014}. We can then express a sample $\vz \sim q(\vz|\vx)$ as,
\begin{align}\label{eq:gaussian_sample}
  \vz &= g_\phi(\vx,\epsilon)
  \\
  &= \mu_\phi(\vx) + R_\phi(\vx)\epsilon,
\end{align}
so that $\vz$ is distributed as multivariate normal with mean $\mu_\phi(\vx)$ and covariance $\Sigma_\phi(\vx) = R_\phi(\vx) {R_\phi(\vx)}\trp$:
\begin{align}\label{eq:general_gaussian_approximate_posterior}
  q(\vz|\vx) = \mathcal{N}\left(\mu_\phi(\vx), \Sigma_\phi(\vx)\right).
\end{align}

It is easy to sample from a Gaussian approximate posterior using \eqref{gaussian_sample}, and the entropy term within \eqref{lower_bound} has a closed form:
\begin{align}
  H(q_\phi(\vz|\vx)) = - \E_{q_\phi(\vz|\vx)} \left[\log q_\phi(\vz|\vx)\right] = \frac{nT}{2}\left(1 + \log(2\pi)\right) + \frac{1}{2} \log \mathrm{det}(\Sigma_\phi(\vx)).
\end{align}

A potential downside of the Gaussian approach is that in our setting $\Sigma \in \R^{nT\times nT}$, and so the number of parameters scales quadratically in $T$. This makes it difficult to manage and learn for large-scale datasets. A simple workaround is to consider $\Sigma$ with a special structure that reduces the effective number of parameters. Possible examples include using diagonal covariance (fully-factorized, or ``mean field'' approximation) \citep{Bishop2006}, or a diagonal covariance matrix plus low-rank matrix (for instance, a sum of outer products) \citep{Rezende2014}.

\subsection{Smoothing Gaussian approximate posterior} \label{sec:smoothing_posterior}

For modeling time-series data, we seek an approximate posterior capable of expressing our strong expectation that the latent variables change smoothly over time. While the Gaussian approximate posterior of~\eqref{general_gaussian_approximate_posterior} can represent arbitrary correlation structure, we propose a Gaussian approximate posterior whose parameterization scales only linearly in $T$. To do so, we borrow from the toolkit of the standard Kalman filter. In an LDS model with Gaussian observations, the posterior is a multivariate Gaussian with a block tri-diagonal inverse covariance. This block-tridiagonal structure results from (and expresses) the conditional independence properties of the LDS prior.

To enable our approximate posterior to express the same correlation structure we parameterize the inverse covariance of~\eqref{general_gaussian_approximate_posterior}, $\inv{\Sigma}$, to be block tri-diagonal. 
Our final posterior takes the form:
\begin{align}\label{eq:general_smoothing_posterior}
  q(\vz | \vx) = \mathcal{N}\left(\mu_\phi(\vx), \inv{\left[R_\phi(\vx) R_\phi(\vx)\trp\right]}\right),
\end{align}
where $\mu_\phi(\vx)$ is the posterior mean and $R_\phi(\vx)$ is a lower block bi-diagonal matrix with $n\times n$ blocks \footnote{A lower block bi-diagonal matrix has only non-zero diagonal and (first) lower-diagonal blocks.}.

\subsubsection{Computation}\label{sec:smoothing_computation}
In this subsection we drop subscripts and functional notation for clarity and refer, for instance, to $\Sigma_\phi(\vx)$ as $\Sigma$. 
We can perform inference efficiently by exploiting the special structure of $\Sigma$. 

While $\Sigma$ is in general a dense matrix, we parameterize $\inv \Sigma$ as block tri-diagonal. Since $\Sigma$ is symmetric, in practice we represent only the diagonal and first block off-diagonal matrices of $\inv \Sigma$. Matrix inversion and sampling may be performed quickly using the Cholesky decomposition\footnote{Block structure is frequently exploited in computation of the Cholesky decomposition; see for instance \cite{Bjorck1996}.}, $\inv \Sigma = RR\trp$. The computation of the lower-triangular Cholesky factor, $R$, is  linear (in both time in space) in the length of the time-series $T$.

We can sample as in \eqref{gaussian_sample}, where now:
\begin{align}\label{eq:smoothed_gaussian_sample}
  \vz = \mu + R\itrp \epsilon.
\end{align}

For an arbitrary matrix $R \in \R^{nT\times nT}$, computation of $R\itrp \epsilon$ scales cubically in dimensionality of the matrix. However, by exploiting the lower-triangular structure of $R$, matrix inversion scales only linearly \citep{Trefethen1997}. The entropy of $q$ is also easy to compute since $\log \mathrm{det}(\Sigma) = -2\log \mathrm{det}(R) = -2\sum_{i=1}^T\log(R_{ii})$.

In short, for learning $\phi$ and $\theta$ we need never explicitly represent any part of $\Sigma$. For data analysis and model comparison, however, it may be useful to compute the covariance $\mathrm{cov}(\vz_t, \vz_{t+1})$. These covariances correspond to the block-diagonal and first block off-diagonals of $\Sigma$, and may also be computed efficiently \citep{Jain2007}.

Our overall approach is closely related to the standard forward-backward algorithm used for instance in Kalman smoothing. However, there is a major technical distinction between its standard use (e.g., in expectation maximization) and our approach: we explicitly differentiate parameters $\phi$ through the matrix factorization $\inv{\Sigma} = R R\trp$. 

\section{Parameterization of the smoothing posterior}\label{sec:parameterization}

While \eqref{general_smoothing_posterior} succinctly states the general mathematical form of the smoothing posterior, the practical performance of the algorithm depends upon the specifics of the parameterization. There are many possible parameterizations, especially since the parameters $\Sigma_\phi(\vx)$ and $\mu_\phi(\vx)$ may be arbitrary functions of observations $\vx$. To illustrate, we discuss two distinct parameterizations. 
We use the notation $P = \mathrm{NN}_{\phi_P}(\vx)$ to indicate that parameter $P$ is defined as a function of inputs $\vx$ through a neural network $\mathrm{NN}_{\phi_P}(\cdot)$ with parameters $\phi_{P}$. The parameters of all networks are incorporated into $\phi$:
\begin{equation}
\phi = \left\{ \phi_{P_1}, \phi_{P_2}, \phi_{P_3}, \dots \right\}.
\end{equation}

\subsection{Diagonal and block off-diagonal parameterization} \label{sec:triple_net}
We can naturally parameterize $\mu_\phi(\vx)$ and $\Sigma_\phi(\vx)$ of \eqref{general_smoothing_posterior} using 3 neural networks. We use one neural network to represent a map $\vx_t \to \mu_t$,
\begin{equation}
  \mu_t = \mathrm{NN}_{\phi_\mu} (\vx_t),
\end{equation}

where $\mu_t$ is a $n\times 1$ segment of $\mu$, and $\mu= \left(\mu_1, \mu_2, \dots, \mu_T\right)$. We can parameterize the block tri-diagonal covariance $\inv{\Sigma_\phi(\vx)}$,
\begin{equation}
\inv{\Sigma_\phi(\vx)} =
    \begin{bmatrix}
        D_0 & B_0\trp \\
        B_0 & D_1 & B_1\trp \\
        & \ddots & \ddots & B_{T-1}\trp\\
        & & B_{T-1} & D_T
    \end{bmatrix},
  \end{equation}
by parmeterizing each of the blocks separately:
\begin{align}
  D_t &= \mathrm{NN}_{\phi_D}(\vx_t)
        \\
  B_t &= \mathrm{NN}_{\phi_B}(\vx_t, \vx_{t-1}).\label{eq:cross_cov_net}
\end{align}

In practice, we found it necessary to enforce the positive-definiteness of the covariance by adding a diagonal matrix $\alpha I$ to $\inv{\Sigma_\phi(\vx)}$, where $\alpha>0$ is a fixed constant. In the experiments, we refer to this parameterization as VILDSblk. 

\subsection{Product-of-Gaussians approximate posterior}\label{sec:lds_rep}
We can also define the approximate posterior through a product of Gaussian factors, $q(\vz|\vx) \propto r_1(\vz|\vx) r_0(\vz)$, where:
\begin{eqnarray}
  r_0(\vz)  &:=& \mathcal{N}(\vz | 0, \mD)\\
  r_1(\vz|\vx)    &:=&  \mathcal{N}(\vz | \mM_\phi(\vx),\mC_\phi(\vx)),
\end{eqnarray}
$\mD$ and $\mC$ are $nT\times nT$ matrices and $\mM$ is a $nT$-dimensional vector. In this set-up, we can view $r_0$ as a prior. In terms of \eqref{general_smoothing_posterior}, the final posterior is then given by:
\begin{align}\label{eq:smooth_posterior_mean}
  \Sigma_\phi(\vx) &= \inv{\left(\inv{\mD} + \inv{\mC}_\phi(\vx) \right)}\\
  \label{eq:smooth_posterior_cov}
  \mu_\phi(\vx)&= \Sigma_\phi(\vx)\inv{\mC}_\phi(\vx) \mM_\phi (\vx).
\end{align}

In order to be a parameterization of the smoothing posterior, \eqref{general_smoothing_posterior}, $\inv{\mD}$ and $\inv{\mC}$ must be block tri-diagonal. The multiplicative interaction between the posterior mean and covariance leads to different performance from the parameterization described in Section \ref{sec:triple_net}. Further, we can choose $\mD$ to initialize the means with a given degree of smoothness, which is not possible in the formulation of Section \ref{sec:triple_net}. In the experiments, we refer to this parameterization as VILDSmult; in Appendix \ref{sec:lds_approx} we describe the specific parameteriation we used for $\inv{\mC}$ and $\inv{\mD}$. 

\section{Experiments}

In the experiments, we refer to the SGVB with the smoothing approximate posterior as VILDS. We refer to SGVB with an approximate posterior independent across time as mean field (MF). The mean field posterior is given by,
\begin{equation}
  \label{eq:mean_field_q}
  q(\vz|\vx)  = \prod_{t=1}^T \mathcal{N}(\vz_t; \mu_t, V_t),
\end{equation}
where $V_t$ is a full $n\times n$ covariance matrix. We optimize all parameters by gradient ascent, using the SGVB approach with $L=1$ to estimate the gradient with~\eqref{sgvb_estimator}. 

For training VILDS and MF, we performed gradient descent on all parameters $\left\{\theta, \phi\right\}$ of the generative model and approximate posterior. We tried several adaptive gradient stochastic optimization methods, including: ADAM \citep{Kingma2014adam}, Adadelta \citep{Zeiler2012}, Adagrad \citep{Duchi2011} and RMSprop \citep{Tieleman2012}. In the experiments we show here, we used Adadelta to learn all parameters. We gradually decreased the base learning rate by a factor of $10$ after a period of $20$ ``epochs'' without an increase of the objective function.

\begin{figure}[t]
\centering
\includegraphics[width=.9\textwidth]{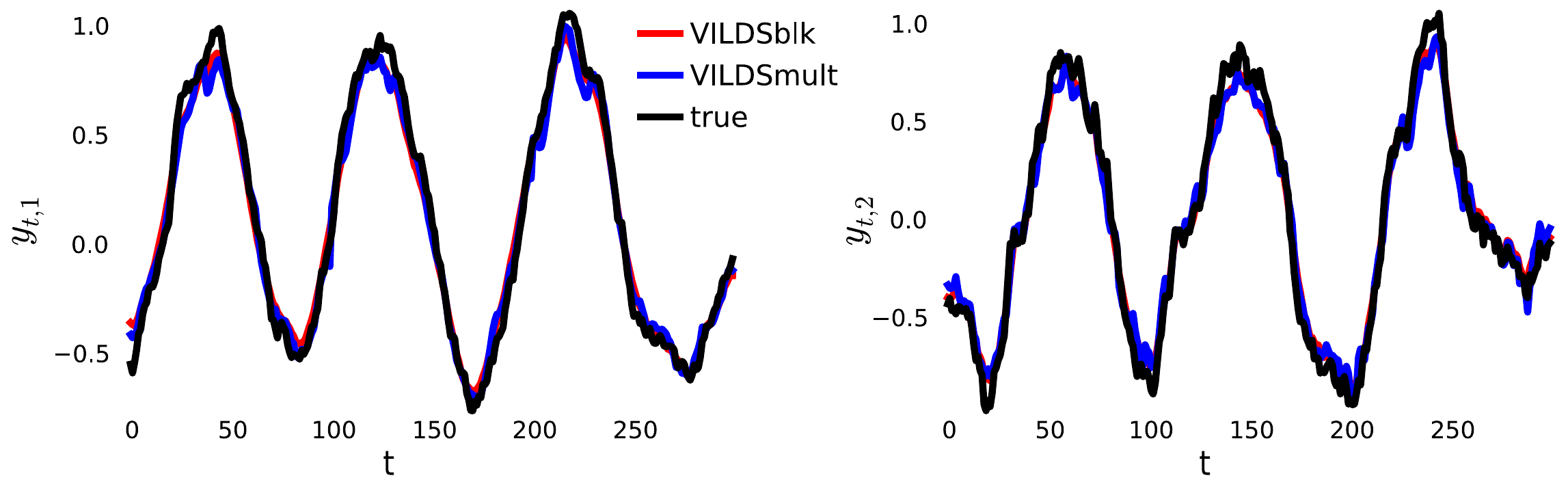}
\caption{Posterior mean inference compared with ground truth. Each panel shows the posterior mean along a dimension of the two-dimensional ($n=2$) state space of a Kalman filter. We show $300$ time-points of the posterior means from a $T=5000$ sample Kalman filter experiment. We fit using VILDS with the parameterization described in Section \ref{sec:triple_net} (VILDSblk) and that described in Section \ref{sec:lds_rep} (VILDSmult). The true posterior means computed using the closed-form Kalman filter equations \textbf{(black)} agree closely with those recovered using VILDSmult \textbf{(blue)} and VILDSblk \textbf{(red)}.}\label{fig:kalman_means}
\end{figure}
\subsection{Kalman filter model}

First, we illustrate the efficacy of our approach by showing that we can recover the analytic posterior of a Kalman filter model. Under a Kalman filter model, the latents are governed by an LDS,
\begin{align}
  \vz_{t}&=\mA\vz_{t-1} + \vepsilon_t, \label{eq:dynamics}
\end{align}
with Gaussian innovation noise with covariance matrix $\mQ$, $\vepsilon_t \sim \mathcal{N}(0, \mQ)$. Observations are coupled to the latents through a loading matrix $\mC$,
\begin{align}
  \vx_t &= \mC \vz_t + \eta_t,
  \label{eq:observation_rate}
\end{align}
and $\eta_t$ are Gaussian noise with diagonal covariance.

We simulated 5000 time-points from a 2-dimensional latent dynamical system model, with 100-dimensional linear observations. We parameterize the VILDS approximate posterior using a 5-layer, dense NN for each of $\mu_\phi(\vx)$ and $R_\phi(\vx)$. We use a rectified-linear nonlinearity between each layer, followed by a linear output layer mapping into the parameterization. We compare both of the approximate posterior parameterizations described in Section \ref{sec:parameterization}. We refer to the parameterization of Section \ref{sec:triple_net} as VILDSblk, and that of Section \ref{sec:lds_rep} as VILDSmult. For both choices of approximate posterior, the VILDS smoothed posterior means (\figref{kalman_means}) show good agreement with the true Kalman filter posterior. The VILDS smoothed posterior variances, $\V[\vz_t]$ and $\mathrm{cov}(\vz_t, \vz_{t-1})$ also show good agreement with the Kalman filter posterior covariance (not shown).

\subsection{Poisson LDS (PLDS)}
\begin{figure}[t]
\centering
  \includegraphics[width=.8\textwidth]{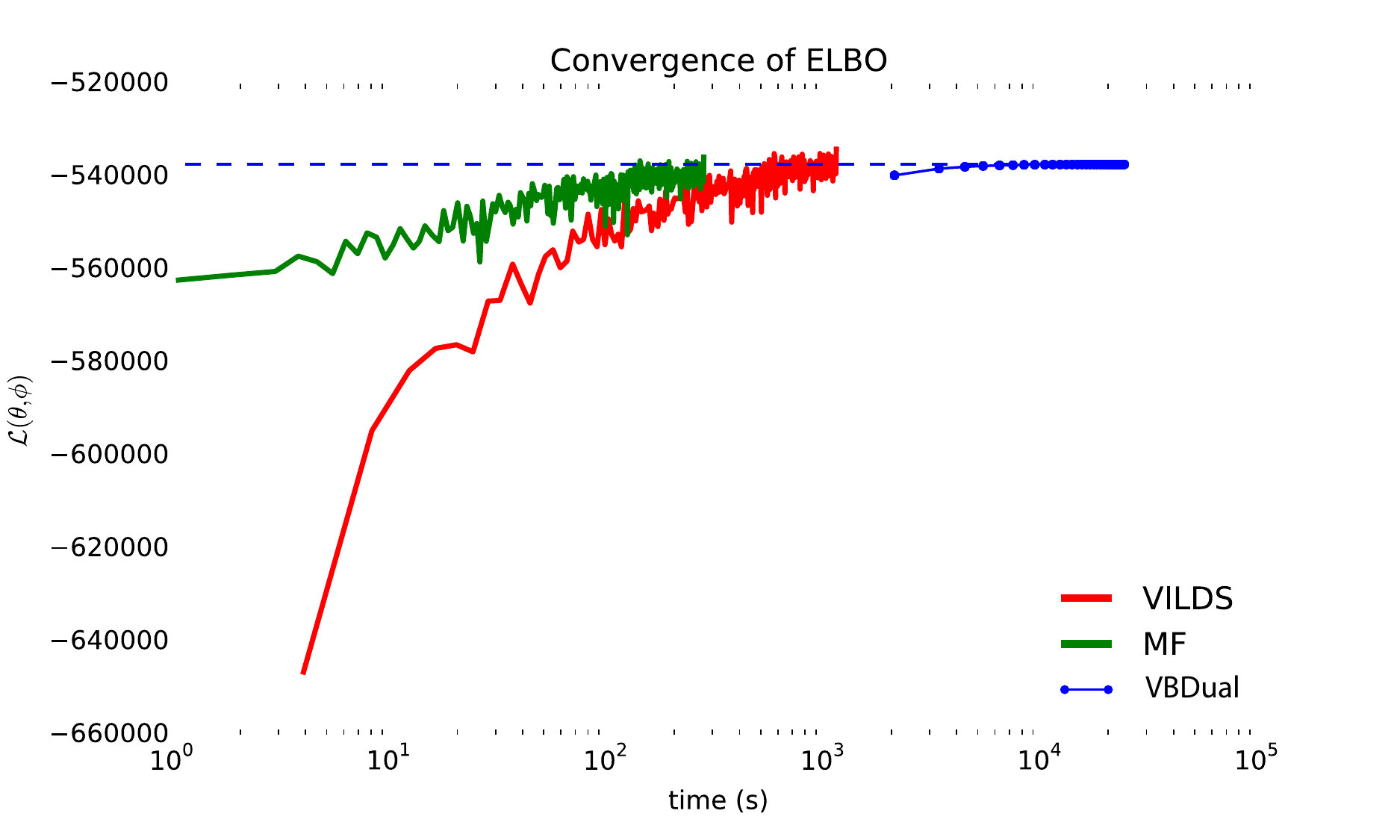}
  \caption{Speed comparison: ELBO convergence vs time (seconds) for VILDS, MF and VBDual. VBDual is fit by an iterative procedure that optimizes a dual-space cost function for each E-step \citep{Kahn2013}. The first VBDual E-step takes many iterations to converge, and causes the long gap from $0$ seconds to the first VBDual datapoint. Subsequent VBDual E-steps are less time-consuming. VILDS and MF are learned by stochastic gradient descent using the adaptive-gradient technique, Adadelta~\citep{Zeiler2012}. Gradients are computed on minibatches of size $100$, and each datapoint is collected after $100$ minibatches (one ``epoch''). Both MF and VILDS were run for $500$ epochs. VILDS achieves the highest ELBO value, followed by MF and then VBDual. VILDS achieves ELBO values comparable to VBDual before VBDual can complete a single EM iteration.}\label{fig:vbem_convergence}
\end{figure}
The Kalman filter/smoother is exact for an LDS with linear-Gaussian observations. A common generalization in the literature is an LDS with non-Gaussian observations. One well-studied example is the Poisson LDS (PLDS). Under this model, the latents are again governed by an LDS,
\begin{align}
  \vz_{t}&=\mA\vz_{t-1} + \vepsilon_t, \label{eq:dynamics}
\end{align}
with Gaussian innovation noise with covariance matrix $\mQ$, $\vepsilon_t \sim \mathcal{N}(0, \mQ)$. Observations are modulated by a log-rate $\vr_t$, which is coupled to the latent state $\vz_t$ via a loading matrix $\mC$,
\begin{align}
  \vr_t &= \mC \vz_t + \vd.
  \label{eq:observation_rate}
\end{align}
The vector $\vd$ is a vector bias term for each element of the response. Given the log-rate $\vr_t$, observations $\vx_t\in \R^m$ are Poisson-distributed,
\begin{align}
  x_{k,t}|\vz_t &\sim \mathrm{Poisson}(\exp(r_{k,t})).
  \label{eq:poiss_obs}
\end{align}
With Poisson observations, the posterior does not have a closed form. Several methods have been proposed for approximate learning and inference in the special case of the PLDS \citep{Buesing2014, Buesing2012, Macke2011}; Laplace approximation is also frequently used \citep{Paninski2010, Fahrmeir1991}. We compare VILDS to the variational Bayes expectation-maximization approach proposed by \cite{Kahn2013}. This VBEM approach uses a full, unconstrained Gaussian as a variational approximate posterior $q_\phi$, and performs EM iterations through a dual-space parameterization. We refer to it by the abbreviation VBDual, to emphasize this dual-space parameterization. We parameterize both the MF and VILDS approximate posteriors using a 5-layer, dense NN for each of $\mu_\phi(\vx)$ and $R_\phi(\vx)$. For VILDS, we use the parameterization of Section \ref{sec:lds_rep}. We use a rectified-linear nonlinearity between each layer, followed by a linear output layer mapping into the parameterization. We simulated $T = 5000$ samples from a PLDS model with $n=2$ latent states and $m=100$ observation dimensions. We initialized all three methods (VILDS, MF and VBDual) using the nuclear-norm minimization methods outlined in \cite{Pfau2013}.

\begin{figure}[t]
\begin{center}
  \includegraphics[width=\textwidth]{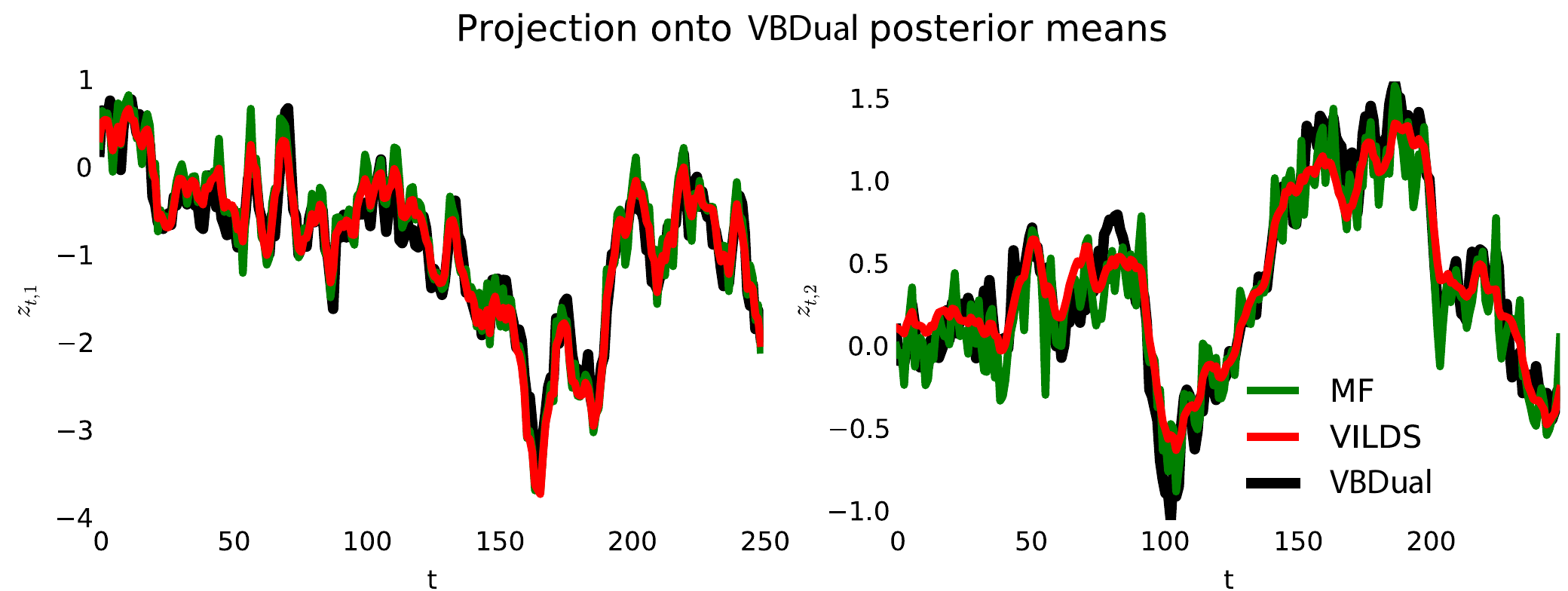}
\end{center}
\caption{Comparison of posterior means learned using the mean field approximate posterior \textbf{(green)}, VILDS \textbf{(red)} and VBDual with an unstructured Gaussian approximate posterior \textbf{(black)}. The model has a rotational invariance, and so we project the MF and VILDS posteriors onto the VBDual posterior means using least squares. The posterior mean trajectories learned by VILDS are visibly smoother than the MF posterior mean trajectories.
}\label{fig:illustrate_smoothing}
\end{figure}

To better illustrate the timecourse of learning, each epoch consisted of only $100$ minibatches, where each minibatch was of size $100$. A single gradient step was taken for each minibatch. We ran both MF and VILDS for a fixed 500 iterations. We find that VILDS reaches a higher ELBO value than either MF or VBDual, and does so before VBDual can complete a single expectation-maximization iteration (see \figref{vbem_convergence}). Further, the posterior means learned by VILDS are smoother than those learned using the MF approximate posterior (see \figref{illustrate_smoothing}).

\subsection{Nonlinear dynamics simulation}
\begin{figure}[t]
\begin{center}
  \includegraphics[width=\textwidth]{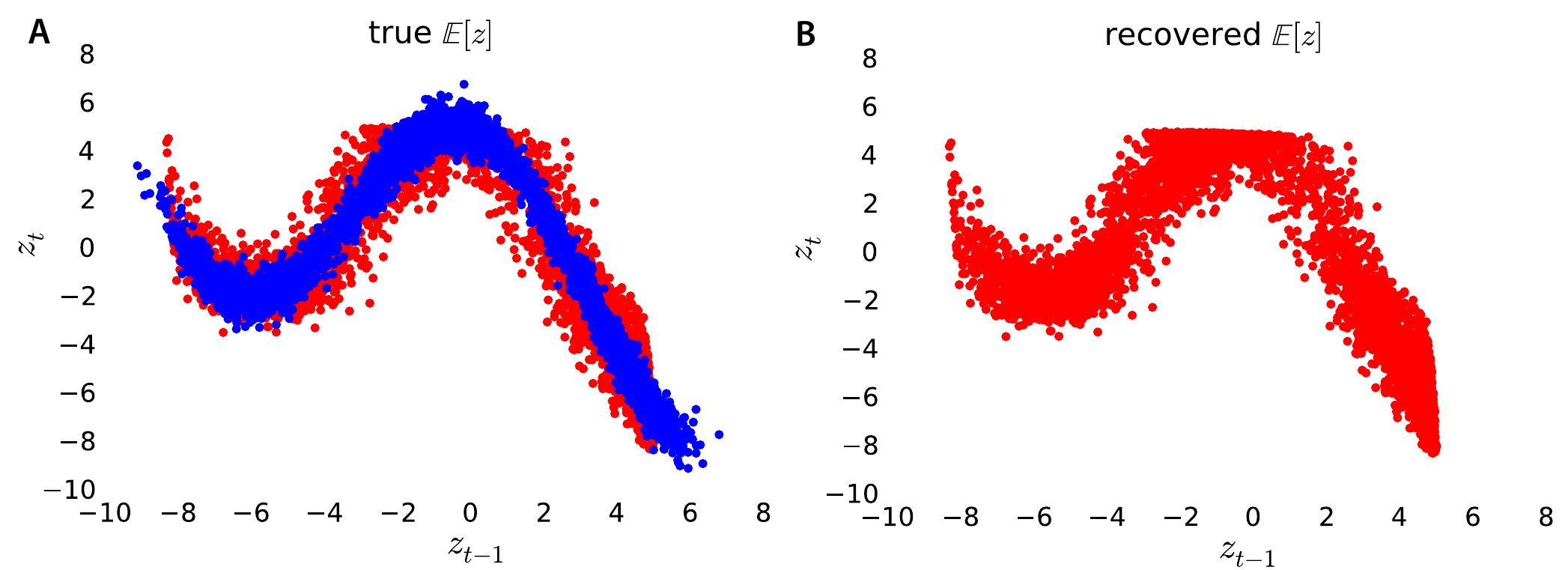}
\end{center}
\caption{True and VILDS-fit posterior means for nonlinear dynamics simulation. VILDS was fit to 5000 samples drawn from the nonlinear dynamical system described in \eqref{nlin_dyn}. Each point in both plots represents a single ordered pair $\left(\E[z_{t}], \E[z_{t-1}]\right)$. \textbf{(A)} We illustrate the nonlinear dynamical relationship between $\vz_t$ and $\vz_{t-1}$; in a linear dynamical system this relationship would be a straight line. In red, we show the posterior means recovered by VILDS.
\textbf{(B)} The VILDS posterior means of \textbf{(A)} plotted alone, for comparison.
}\label{fig:nonlinear_dynamics}
\end{figure}

VILDS can perform approximate posterior inference for nonlinear, non-Gaussian generative models. To illustrate, we simulated $5000$ samples from  a toy one-dimensional nonlinear dynamical model given by:
\begin{align}\label{eq:nlin_dyn}
  \vz_t &= -\frac{1}{2}\vz_{t-1} + 5 \cos(.5\vz_{t-1}) + .5 \epsilon_t
  \\
  \vx_t &= \frac{1}{2}\vz_{t} + .5 \eta_t,
\end{align}
where $\epsilon_t$ and $\eta_t$ are each iid $\mathcal{N}(0,1)$ random variables. For the approximate posterior, we parameterized both the $\mu$ and $R$ using 8-layer networks where each layer has only a single unit, and rectified-linear nonlinearity. We use the LDS-inspired parameterization of Section \ref{sec:lds_rep}. As shown in \figref{nonlinear_dynamics}, VILDS is capable of recovering the nonlinear relationship in the state space. For these experiments, we held the generative model parameters $\theta$ fixed and learned only $\phi$. 

\section{Conclusion}
We proposed a Gaussian variational approximate posterior with block tri-diagonal covariance structure capable of expressing ``smoothed'' trajectories of a time-series posterior. Exploiting the block tri-diagonal covariance structure, inference scales only linearly (in both time and space complexity) in the length $T$ of a time-series. Using the SGVB approach to variational inference, we can perform approximate inference for a wide class of latent variable generative models. 

Despite the generality of the inference algorithm, the approach is limited by the Gaussian approximate posterior: most latent variable time-series generative models have non-Gaussian posteriors. One possible route forward are the methods of \cite{Rezende2015} and \cite{Dinh2014}, which permit learning and inference using a non-Gaussian approximate posterior within the SGVB framework.

We implemented all methods in Python using Theano with the Lasagne library \citep{Bastien2012, Bergstra2010}. We plan to release the source code on Github shortly.

As we were preparing this manuscript we became aware of \cite{Krishnan2015}, which studies a closely-related (but distinct) method. In future work, we will plan to perform detailed comparisons between the methods. 

\subsubsection*{Acknowledgments}
Funding for this research was provided by DARPA N66001-15-C-4032, Google Faculty Research award, and ONR N00014-14-1-0243 (LP);
Simons Global Brain Research Award 325171 (LP and JC); Sloan Research Fellowship (JPC). We thank David Carlson and Megan McKinney, Esq. for useful comments on the manuscript, and Gabriel Synnaeve for making his Python code publicly available. 

\bibliographystyle{iclr2016_conference}
\bibliography{\mybibfile}

\appendix

\section{Smoothing approximate posterior with explicit forward/backward}\label{sec:lds_approx}
The posterior mean and covariances may be computed by the standard Kalman forward-backward algorithm. To see this, we can write the posterior in matrix form as the product of two Gaussians. We have
\begin{eqnarray}
  r_0(\vz)  &:=&   \mathcal{N}(\vz | 0, \mD) 
  \\
  r_1(\vz|\vx)    &:=& \mathcal{N}(\vz | \mM_\phi(\vx),\mC_\phi(\vx)),
\end{eqnarray}
where,
\begin{equation}
    \mQ = \Isz{T}{T}  \otimes Q,
\quad
    \mA =
    \begin{bmatrix}
        0 \\
        1 & 0 \\
        & \ddots & \ddots \\
        & & 1 & 0
    \end{bmatrix}
    \otimes A, \quad \mD = (I-\mA)\itrp\mQ\inv{(I-\mA)},
\end{equation}

where the positive-definite matrix $Q$ is a covariance matrix, analogous to the ``innovation noise'' in the standard Kalman filter, $n\times n$ matrix $A$ is a linear dynamics matrix\footnote{For stable dynamics, we assume that the eigenvalues of $A$ have magnitude less than one; in the examples we considered, we did not need to enforce this constraint.}. 

We can then re-write the approximate posterior $q$ as the product $q(\vz|\vx) \propto r_0(\vz) r_1(\vz)$. By the standard product-of-normal-densities identity, $q(\vz|\vx)$ also has a multivariate normal distribution $q = \mathcal{N}(\mu_{\phi}(\vx), \Sigma_{\phi}(\vx))$, where  $\mu$ and $\Sigma$ are given by \eqref{smooth_posterior_mean} (which we repeat here):
\begin{align}
  \Sigma_\phi(\vx) &= \inv{\left(\inv{\mD} + \inv{\mC}_\phi(\vx) \right)}\\
  \label{eq:smooth_posterior_cov}
  \mu_\phi(\vx)&= \inv{\left(\inv{\mD} + \inv{\mC}_\phi(\vx) \right)} \inv{\mC}_\phi(\vx) \mM_\phi (\vx).
\end{align}

Computation proceeds just as in Section~\ref{sec:smoothing_computation}, except that now the computation of \eqref{smooth_posterior_mean} takes the form,
\begin{align}
  \mu = \inv{(RR\trp)}\inv\mC M = R\itrp (\inv R (\inv \mC M))
\end{align}
which may be computed efficiently by exploiting the block bi-diagonal structure of $R$.

\end{document}